\documentclass{article}

\usepackage[margin=3.2cm]{geometry}

\usepackage{parskip}

\usepackage{hyperref}
\usepackage{url}
\usepackage{graphicx}
\usepackage{mathtools,amsthm,amssymb}
\usepackage{xspace}
\usepackage{colortbl, booktabs} \usepackage{subcaption}
\usepackage{tabularx}
\usepackage{array}
\usepackage{makecell}
\usepackage{enumitem}
\usepackage{overpic}

\usepackage{natbib}
\usepackage{xcolor}
\definecolor{mydarkblue}{rgb}{0,0.08,0.45}
\hypersetup{     colorlinks=true,
    linkcolor=mydarkblue,
    citecolor=mydarkblue,
    filecolor=mydarkblue,
    urlcolor=mydarkblue,
}

\usepackage[p,osf]{cochineal}
\usepackage[scale=.95,type1]{cabin}
\usepackage[cochineal,bigdelims,cmintegrals,vvarbb]{newtxmath}
\usepackage[zerostyle=c,scaled=.94]{newtxtt}
\usepackage[cal=boondoxo]{mathalfa}
\usepackage[scale=0.9]{inconsolata}

\usepackage[protrusion=true,expansion=true]{microtype}

\theoremstyle{plain}

\theoremstyle{definition}

\theoremstyle{remark}

\newcommand{\papertitle}{Pointer Value Retrieval: A new benchmark for understanding the limits of neural network generalization}
\title{\papertitle}
\date{}

\usepackage[auth-lg,affil-sl]{authblk}

\newcommand*\samethanks[1][\value{footnote}]{\footnotemark[#1]}
\author[1]{Chiyuan Zhang\samethanks[1]}
\author[1]{Maithra Raghu\thanks{Equal Contribution}}
\author[2]{Jon Kleinberg}
\author[1]{Samy Bengio\thanks{Currently at Apple}}

\affil[1]{Google Research}
\affil[2]{Cornell University}

\usepackage{nicefrac}       \usepackage{microtype}      \usepackage{xcolor}         \usepackage{enumitem}
\usepackage{grffile}
\usepackage[export]{adjustbox}
\usepackage{overpic}
\usepackage{sidecap}

\newcommand{\nsmath}{\mathbf{\operatorname{NS}}}
\newcommand{\modsum}{\textit{mod\_sum}\xspace}
\newcommand{\mlp}{\texttt{MLP}\xspace}
\newcommand{\mlptwox}{\texttt{MLP 2x}\xspace}
\newcommand{\transformer}{\texttt{Transformer}\xspace}
\newcommand{\mlpmixer}{\texttt{MLP-Mixer}\xspace}
\newcommand{\resnet}{\texttt{ResNet18}\xspace}
\newcommand{\vgg}{\texttt{VGG11bn}\xspace}

\usepackage{xspace}

\begin{document}
\maketitle

\begin{abstract}
Central to the success of artificial neural networks is their ability to generalize. But does neural network generalization primarily rely on seeing highly similar training examples (\textit{memorization})? Or are neural networks capable of human-intelligence styled \textit{reasoning}, and if so, to what extent? These remain fundamental open questions on artificial neural networks. In this paper, as steps towards answering these questions, we introduce a new benchmark, \textit{Pointer Value Retrieval} (PVR) to study the limits of neural network reasoning. The PVR suite of tasks is based on reasoning about \textit{indirection}, a hallmark of human intelligence, where a first stage (task) contains instructions for solving a second stage (task). In PVR, this is done by having one part of the task input act as a \textit{pointer}, giving instructions on a different input location, which forms the output. We show this simple rule can be applied to create a diverse set of tasks across different input modalities and configurations. Importantly, this use of indirection enables systematically varying task difficulty through distribution shifts and increasing functional complexity. We conduct a detailed empirical study of different PVR tasks, discovering large variations in performance across dataset sizes, neural network architectures and task complexity. Further, by incorporating distribution shift and increased functional complexity, we develop nuanced tests for reasoning, revealing subtle failures and surprising successes, suggesting many promising directions of exploration on this benchmark.
\end{abstract}

\section{Introduction}
Neural networks have demonstrated extraordinary successes across a variety of domains \citep{brown2020language, jumper2021highly, han2020streaming, ettinger2021large, radford2021learning}, through their ability to generalize to unseen data. Despite these successes, there remain fundamental open questions on the inner workings of neural network generalization \citep[e.g.][]{zhang2016understanding,Neyshabur2018-zn,khandelwal2019generalization, feldman2020neural}. Specifically, how much do neural networks rely on memorization --- seeing highly similar training examples --- and to what extent are they capable of human-like reasoning?

These questions are of central importance in assessing the limitations of artificial neural networks, a topic that has gained pressing urgency as these models are applied to high-stakes settings such as medical diagnosis, which require robust reasoning capabilities \citep{winkler2019association, oakden2020hidden}.

An active line of research in cognitive and computational neuroscience has shown that one key component of human-like reasoning is the ability to process \textit{indirection} \citep{kriete2013indirection, noelle2017indirection, buchweitz2013bilingual, hayworth2018thalamic, muller2016model}. Indirection comprises a multistage reasoning process, where the first stage provides \textit{instructions} on performing subsequent tasks. The (human) capability to understand and follow these variable instructions enables going beyond direct experiences and adapting to unseen circumstances. This ability is linked to the function of the prefrontal cortex, with an analogue of pointer referencing in computer systems \citep{kriete2013indirection, russin2020deep, alexander2015hierarchical}.

Informed by these findings, we introduce a new family of benchmark tasks \textit{Pointer Value Retrieval} (PVR), that provide a rich testbed for understanding the limits of artificial neural network reasoning. PVR tasks vary in input types and complexity, but all make fundamental use of indirection through a \textit{pointer rule}: one part of the input acts as a \textit{pointer}, providing instructions on a specific input location, which is processed to form the output. Crucially, this use of indirection allows PVR task complexity to be systematically varied through increasing functional complexity and adding distribution shift, allowing for nuanced tests revealing insights into neural network generalization. Specifically, our contributions are:
\begin{itemize}[leftmargin=1em,itemsep=0em,topsep=0em]
    \item We introduce Pointer Value Retrieval (PVR), a rich new benchmark for studying the limits artificial neural network reasoning. PVR tasks encapsulate indirection through a pointer rule involving input positions and values. We design PVR tasks across different data modalities and structures, which combine representation learning along with reasoning. 
    \item Indirection and the pointer rule lead to natural ways to systematically vary complexity of PVR tasks through distribution shift and increasing functional complexity. 
    \item We reveal connections between our definitions of functional complexity and noise sensitivity in Boolean functions, and highlight a subtle, systematic generalization failure on PVR tasks with distribution shift.
    \item We perform a detailed empirical exploration of PVR tasks, finding significant performance variations across MLPs, Transformers and the recent MLP-Mixer. We identify a relationship between training dataset size and successful generalization in high complexity PVR tasks.
    \item Such high complexity PVR tasks exhibit instability in learning. We investigate this, discovering differences between fast learners and slow learners. A further analysis with a nuanced PVR task provides a partially successful demonstration of neural network reasoning, and we discuss ramifications for future exploration.  
                        \end{itemize}

\section{Related Work}
\textbf{Memorization.} One puzzle in understanding deep learning is that large overparameterized neural networks with enough capacity to memorize~\citep{zhang2016understanding} or interpolate~\citep{belkin2021fit} the entire training set generalize well in real world data. Previous papers focused on characterizing the generalization bounds of overparameterized models~\citep[e.g.][]{Neyshabur2018-zn,Belkin2018-rf}. Explicit study of memorization were also carried out from different perspectives such as invariances \citep{parascandolo2020learning} generalization~\citep{feldman2020neural} and privacy~\citep{carlini2021extracting}. In this paper, we also consider the overparameterized setting where the neural networks could fit the entire training set, and characterize the compositional generalization and reasoning power in tasks with varying functional complexity.

\textbf{Reasoning.} To precisely control the task complexity, we formulated a Pointer Value Retrieval (PVR) framework. Neural network generalization and reasoning has been studied with several approaches/tasks distinct from our PVR framework. One active direction has looked at the capability of models to learn visual tasks in a few-shot setting \citep{snell2017prototypical, triantafillou2019meta}, on benchmarks such as Omniglot \citep{lake2011one}, and Bongard-LOGO \citep{nie2020bongard}. Our focus is not on few-shot, but faithfully learning the reasoning rule. Previous work has looked at learning rules in visual question answering \citep{Bahdanau2019-iy, johnson2017clevr}, drawing on classical work studying generalization  \citep{fodor1988connectionism}. These rules are often more complex than the simple pointer-value formulation. Other work with varied rules has looked at mathematical reasoning, both in the form of non-verbal, visual puzzles \citep{barrett2018measuring, zhuo2020solving} as well as more symbolic mathematical inputs \citep{saxton2019analysing, russin2021compositional}, with related efforts studied in RL \citep{raghu2018can, zhang2016understanding, oh2017zero}. There are also tasks exploring verbal reasoning abilities \citep{lake2018generalization, nye2020learning}, which are different from our setting with systematic difficulty variation. Pertinent to these are also directions examining sequence-to-sequence learning and more generally compositional generalization \citep{lake2018generalization,Keysers2020-bd, keysers2019measuring, livska2018memorize, ruis2020benchmark}. Our work is related to this but more specifically focuses on indirection. Another line of literature design neural networks with dedicated symbolic components to improve systematic generalization~\citep{cai2017making,chen2020compositional}. We instead focus on studying general architectures and analyze whether generalization behavior is learned from the data.

\textbf{Indirection.} Our task formulation is inspired by the studies of indirection in neuroscience and cognitive science about human reasoning~\citep{kriete2013indirection, noelle2017indirection, buchweitz2013bilingual, hayworth2018thalamic, muller2016model}. It is shown that indirection provides biologically plausible foundation to variable bindings and symbol processing, which are considered crucial components for human reasoning and systematicity.

\section{Pointer Value Retrieval}
Pointer Value Retrieval (PVR) is a new family of tasks for studying the reasoning capabilities of artificial neural networks, particularly their ability to process \textit{indirection}. Indirection is a multistage reasoning process where the first stage provides \emph{instructions} on subsequent tasks, and it forms a key part of human reasoning \citep{kriete2013indirection}. PVR tasks encapsulate indirection through a \textit{pointer rule}: 
\begin{itemize}[leftmargin=1em,itemsep=0.15em,topsep=0em]
    \item A specific position of the input acts as a \textit{pointer}. (In practice we use the leftmost entry.)
    \item The value of the pointer entry provides instruction on which other position(s) of the input to look at.
    \item The value(s) of the input at these other position(s) are aggregated to produce the final output.
\end{itemize}
Thus PVR tasks are a two stage reasoning process with the pointer providing (variable) instruction on on how to find the correct values to compute the output. Perfectly solving the PVR tasks requires an artificial neural network to learn the role of the pointer, interpret its instruction by reading its value, and execute that instruction by reading and aggregating input values at the specified positions. 

Note that this task definition is agnostic to the type of input. In fact, the input modality can be chosen to introduce (or avoid) representation learning along with reasoning. Concretely, below we introduce (i) a token sequence based PVR task (that reduces the complexity of the data representation to focus on reasoning), and (ii) two image-based PVR tasks (that introduce additional complexity in visual representation learning to appropriately interpret input values). 

\subsection{Token Sequence Based Pointer Value Retrieval}
\label{subsec:PVR-vectorized-task}

\begin{figure}
\centering
\includegraphics[width=.45\columnwidth]{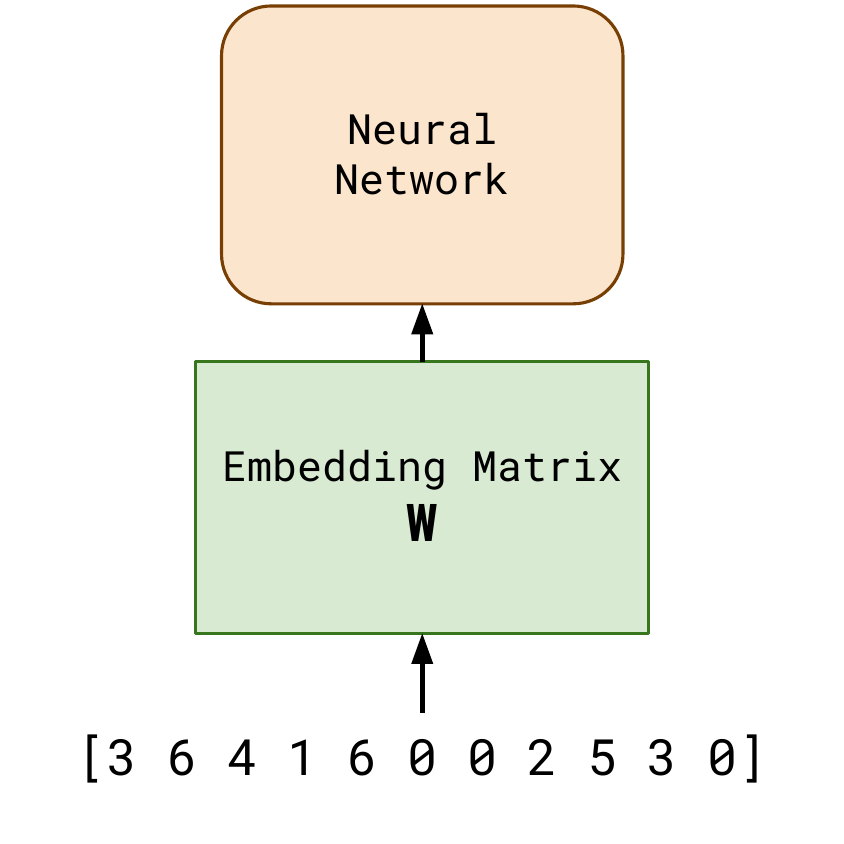}
\vskip-12pt
\caption{\small \textbf{Example of token sequence based PVR task.} The sequence consists of $11$ tokens of integer values between $0-9$. The pointer is given by the left most token, and its value instructs us on the number of positions left we must look (starting from the second most left token.) In the figure, the pointer of value $3$ directs us to the fifth position, of value $6$. In the simplest version of the task, this value directly gives the label. When training on this task, tokens are first embedded with a learnable embedding matrix.}
\label{fig:vectorized-data-examples}
\end{figure}
The simplest PVR task takes in a \textit{sequence} of \textit{tokens} as input. E.g. the input is a sequence of numbers. We take the left most token (number) to be the pointer, and its value determines which input positions to examine next to get the output. In the most basic version of this task, the input is a sequence of $n+1$ numbers, each of which can take values between $0$ and $n$. A pointer with value $m$ instructs us to look at the $m+2$ position from the left, and read out the value as the output. For example, in Figure \ref{fig:vectorized-data-examples}, the pointer directs us to the fifth position from the left, of value $6$, which gives the output. When training on this task, we embed each token with an learnable embedding matrix before passing it as input to the neural network.

\subsection{Visual Pointer Value Retrieval}

\begin{figure}
\centering
\hspace*{-10mm} \includegraphics[width=1.2\columnwidth]{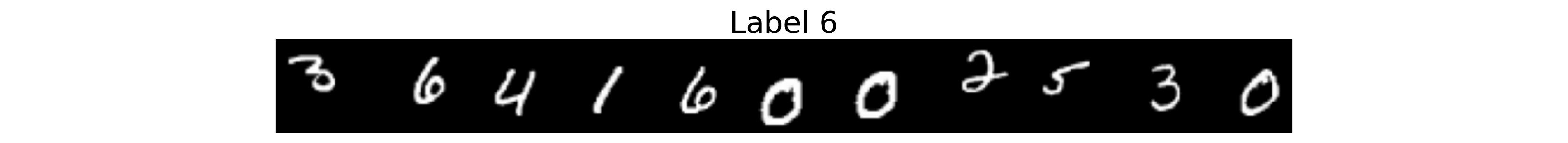}
\vskip-8pt
\caption{\textbf{Example of Image Sequence PVR.} Similar to Figure \ref{fig:vectorized-data-examples}, this task takes in a sequence, but with type of images tokens. The figure shows a simple example of a sequence of digit images. The length of the sequence and mix of image datasets might be varied to change task complexity. Embedding methods, fully convolutional or Transformer architectures can be applied to this task.}
\label{fig:visual-data-examples-sequential}
\end{figure}
To add complexity of accurate representation learning to the reasoning task, we look at visual PVR tasks. These tasks take image inputs and require reasoning (for the pointer rule) and visual understanding (to interpret the values). One obvious visual PVR task is to consider a sequence of \textit{images} instead of a sequence of abstract tokens. A simple example is given in Figure \ref{fig:visual-data-examples-sequential}, with a sequence of digit images. For more complex versions, we might vary the image dataset or the length of the sequencet. We call this image sequence PVR, and similar embedding based (or even fully convolutional or Transformer based) methods can be applied. 

\begin{figure}
\centering
\includegraphics[width=0.8\columnwidth]{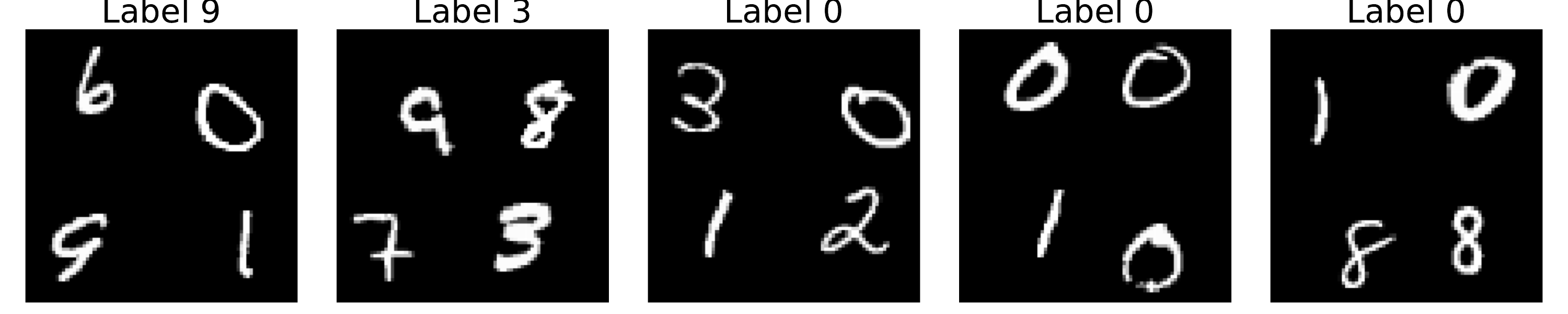} \\
\vspace*{2mm}
\includegraphics[width=0.8\columnwidth]{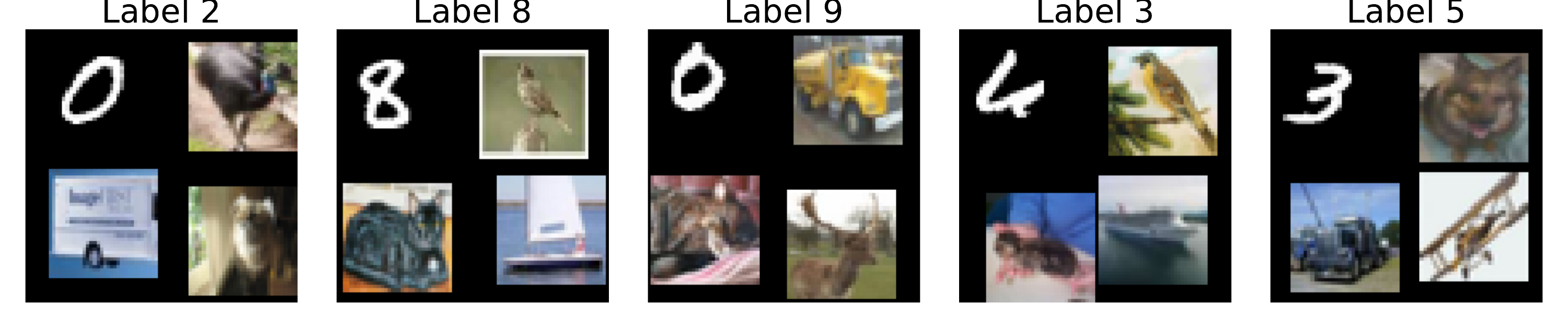}
\vskip-8pt
\caption{\small \textbf{Examples of $2\times 2$ Block Style Image PVR.}  We define the top left image to be the pointer, which defines which one of the remaining three images to examine, and the class of that image is the value. Concretely, we take pointer digits in $0-3$ to indicate the upper right position, $4-6$ the lower left position, and $7-9$ the bottom right position. The image at that position determines the label by using the digit value for top row of MNIST examples, and class label for bottom row of CIFAR-10 examples. Variations of Block Style Visual PVR may use different image datasets, vary the pointer location/instructions, and use a larger $n\times n$ grid.}
\label{fig:visual-data-examples-block-style}
\end{figure}

A more natural configuration for visual PVR is to consider an $n$ x $n$ \textit{block} of images.  One of the $n^2$ constituent images is the pointer, with instructions on which of the other $n^2 - 1$ images must be viewed and aggregated. In the simplest version, we can consider a $2$x$2$ block of images, where we define the top left image to be the pointer. The advantage of square shaped blocks over image sequences is that the state-of-the-art neural network architectures for image recognition typically take square images. Figure \ref{fig:visual-data-examples-block-style} shows some examples with different datasets. In these examples the pointer is a digit image, and the digit value provides instruction on which of the remaining three images determines the output. The most basic version has the output being the class of a single image referenced by the pointer. 

\section{Varying Difficulty of PVR Tasks}
\vspace{-0.5em}
The use of indirection via the pointer rule in the PVR tasks makes it natural to systematically vary their difficulty --- a highly desirable attribute for a benchmark designed to study the limits of neural network reasoning. Below we introduce two different methods to change task complexity: increasing functional complexity and making distribution shift.

\subsection{Varying Difficulty with Functional Complexity}
\label{subsec:PVR-functional-complexity}
\vspace{-0.4em}
\begin{figure}
\centering
\includegraphics[width=.7\columnwidth]{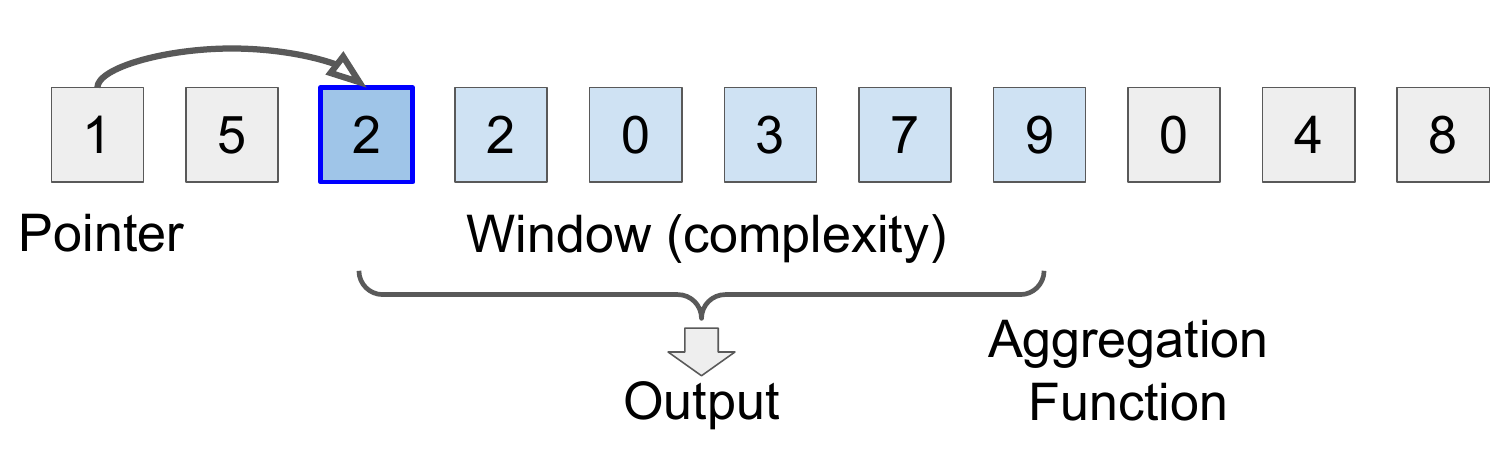}
\vskip-8pt
\caption{\textbf{Increasing functional complexity of PVR tasks.} Instead of the pointer indicating a single position which becomes the value (and output), the pointer instead indicates a \textit{window} of values, which are aggregated to form the final output. The size of the window can be determined by a complexity parameter. The schematic has window size $6$.}
\label{fig:functional-complexity}
\end{figure}

In the simple examples of the token sequence and visual PVR tasks in Figures \ref{fig:vectorized-data-examples}, \ref{fig:visual-data-examples-sequential}, \ref{fig:visual-data-examples-block-style}, the pointer has indicated a specific position, with the value of that position directly providing the output. Instead, the pointer can reference a \textit{window} of values, which are aggregated to form the final output. The size of the window is determined by a fixed, complexity parameter, with window size $1$ corresponding to the simplest task of directly outputting a single position value. Figure \ref{fig:functional-complexity} shows a schematic of this.

When working with numerical sequences, there are several natural aggregation functions, such as (i) \textbf{\modsum} (sum all values in the window and compute remainder mod $10$) (ii) \textbf{\textit{median}} (median of window values) (iii) \textbf{\textit{maj-vote}} (mode of window values) (iv) \textbf{\textit{min/max}} of window values.

But how does introducing a window of positions to aggregate increase task complexity? We can intuitively understand this by observing that the total number of positions used to determine the output has increased, so a small difference in input can cause a large difference in outputs. Specifically, even if an unseen input is very similar to one in the training data, the correct output for it might differ significantly. Thus, an artificial neural network solving this task cannot easily rely on memorization or nearest neighbors, and is pushed to reason about indirection. This intuition on the effect of window size can also be made more formal through measuring \textit{noise sensitivity}.

\begin{figure}
    \centering
    \begin{overpic}[width=.53\linewidth]{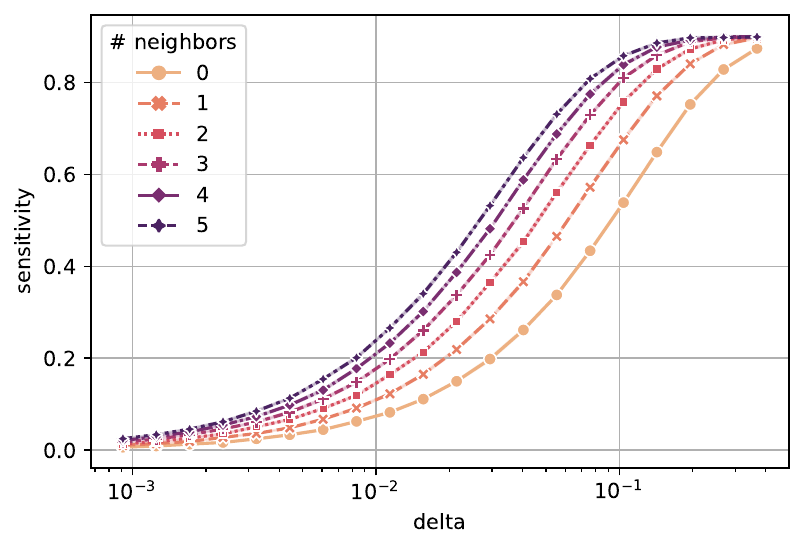}\put(1,5){\textbf{(a)}}\end{overpic}
    \begin{overpic}[width=.46\linewidth]{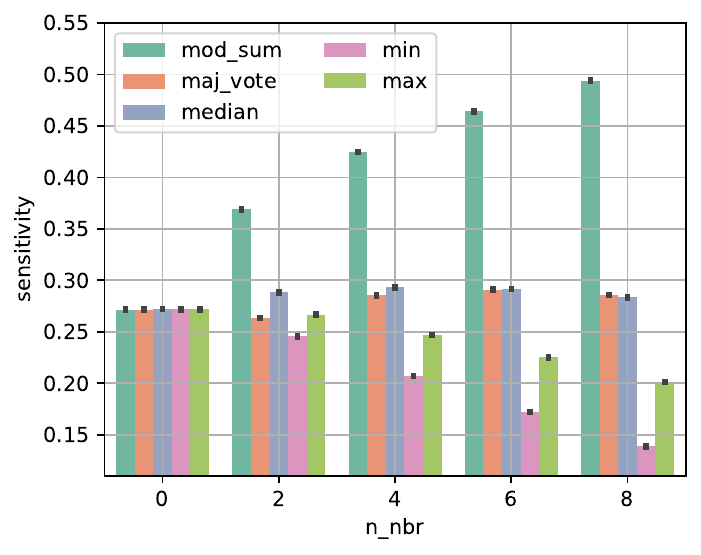}\put(1,5){\textbf{(b)}}\end{overpic}
    \vskip-8pt
    \caption{\small \textbf{Estimated noise sensitivity}. We estimate the noise sensitivity of token sequence based PVR tasks by sampling 10,000 random uniform bit sequences, and report the average estimation over 10 runs. \textbf{(a)} shows $\nsmath_\delta[f]$ for $\delta\in[e^{-7},e^{-1}]$, for the target functions $f$ with \modsum aggregation over different neighborhood sizes. \textbf{(b)} shows the average noise sensitivity over the same value range of $\delta$ across a range of different aggregation choices.}
    \label{fig:vec-noise-sensitivity}
\end{figure}

\begin{figure*}
\centering
    \includegraphics[width=.98\linewidth]{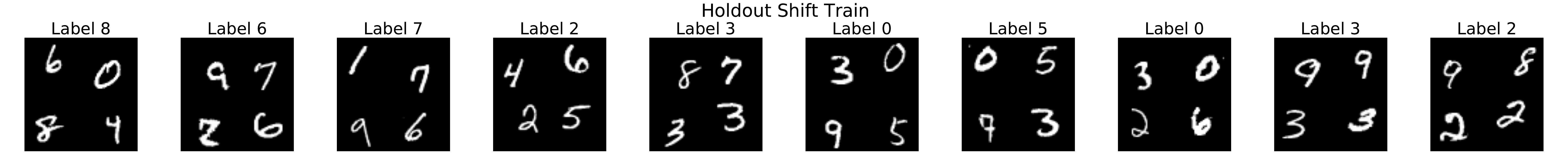} \\\vskip5pt
    \includegraphics[width=.98\linewidth]{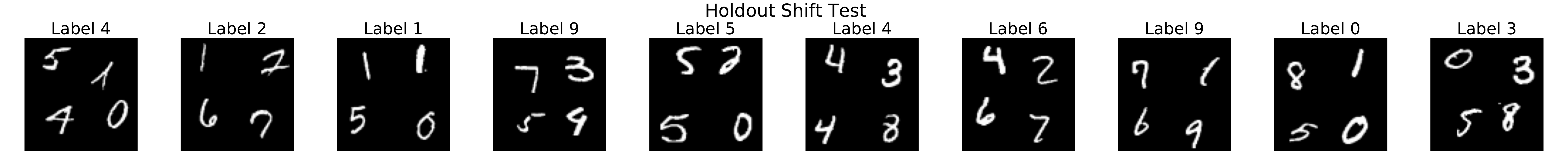}
    \vskip-8pt
\caption{\small \textbf{An example of Holdout (distribution) shift in the $2\times 2$ Block Style PVR task.} We introduce distribution shift by holding out some digits at some positions during training. Specifically in the train set, digits $1-3$ never appear in the top right, $4-6$ never appear in bottom left, $7-9, 0$ never appear in bottom right (top row). At test time, we can test on the \textit{holdout shift}, exactly the examples that were held out at training (bottom row).}
\label{fig:visual-dshift-schematic}
\end{figure*}

\textbf{Measuring complexity with noise sensitivity.}
We apply \emph{noise sensitivity}, used to determine complexity of boolean functions \citep[][Proposition~3.3]{o2014analysis}, to quantify the complexity of our tasks. For a function $f$ taking a boolean sequences as inputs, the noise sensitivity is computed as
\begin{equation}
    \text{NS}_\delta[f] = \mathbb{E}^\delta_{(\mathsf{x},\mathsf{y})}[\mathbf{1}[f(\mathsf{x})\neq f(\mathsf{y})]]
\end{equation}
where the expectation $\mathbb{E}^\delta_{(\mathsf{x},\mathsf{y})}$ is taken by sampling $\mathsf{x}$ as uniformly random bit sequences, and $\mathsf{y}$ by flipping each bit of $\mathsf{x}$ independently with probability $\delta$.
Intuitively, the noise sensitivity measures how sensitive the outcome of $f$ is to random perturbations of the inputs. Figure~\ref{fig:vec-noise-sensitivity}a plots $\text{NS}_\delta[f]$ with varying $\delta$s. The results are consistent with our intuition that target functions with larger window sizes are more complex. In fact, the choice of the \modsum label aggregation is also important. As shown in Figure~\ref{fig:vec-noise-sensitivity}b, other common aggregations such as majority voting don't generate a clean sequence of tasks with increasing difficulties. For \texttt{min} and \texttt{max} aggregation, the tasks actually become easier as window size increases. Additional details are in the Appendix.

\subsection{Varying Difficulty with Distribution Shift}
\vspace{-0.5em}
\label{subsec:vary-difficulty-dshift}
Another natural method to vary difficulty in PVR tasks arises is by introducing \textit{Holdout} distribution shift. As PVR requires reasoning about input positions \textit{and} values, we can holdout all examples with certain values in certain positions at training, and at test time, \textit{only} evaluate on these examples. This can be done while ensuring that all possible pointer instructions and values still occur in the training data. So (i) as there is a significant Holdout shift between train and test, memorization and nearest neighbors cannot be relied on, but (ii) the neural network can still generalize, provided it learns about different positions, values and the pointer rule.

\begin{figure}
\centering
\includegraphics[width=0.6\columnwidth]{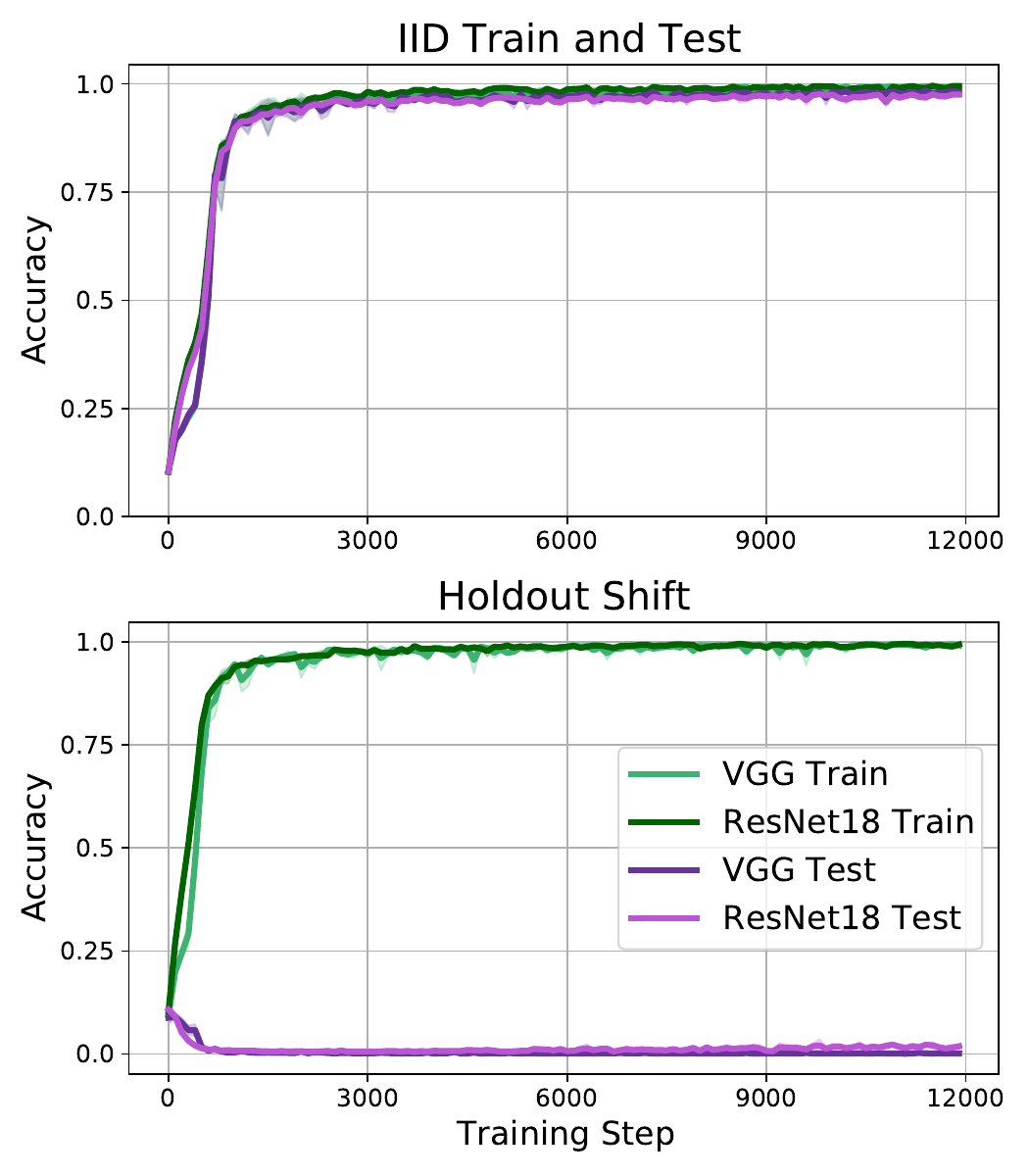}
\vskip-8pt
\caption{\small \textbf{Training and test performance of neural networks on $2\times 2$ Block style visual PVR in IID and Holdout shift settings.} The high training performance across both models and settings contrasted with the very poor generalization in Holdout shift suggests the neural networks are learning by nearest neighbors styled methods, which fail on the Holdout shift. This systematic failure (convergence to \textit{below} random) is further explored in Figure \ref{fig:logits-iid-dshift}.}
\label{fig:visual-iid-holdout}
\end{figure}

A concrete example of this Holdout distribution shift is shown in Figure \ref{fig:visual-dshift-schematic}. We take the $2$ x $2$ block style visual PVR task, and in the training set, make sure digits $1-3$ never appear in the top right, digits $4-6$ never appear in the bottom left and digits $7-9,0$ never appear in the bottom right. The test set consists only of these held out examples.

To understand the Holdout shift and its effects on neural network learning, we conduct some preliminary experiments exploring performance on PVR tasks with Holdout shift. Taking the $2\times 2$ Block Style Visual PVR task, we train two convolutional networks, VGG \citep{simonyan2014very} and ResNet \citep{he2016deep} on an IID version of the task, and the Holdout Shift version defined in Figure \ref{fig:visual-dshift-schematic}. The results, shown in Figure \ref{fig:visual-iid-holdout} are striking, and provide several insights on the mechanisms of neural network learning.

By looking at the top pane, we observe that in the IID setting, both neural network architectures train and generalize to $100\%$ performance. However, in the bottom pane, with Holdout Shift, we see that both models completely fail to generalize, \textit{but} still achieve $100\%$ training accuracy. This indicates that in these settings, the neural networks are achieving high training accuracy not by learning to reason, but by methods such as nearest neighbors, mapping unseen test examples to the most similar training examples. This strategy works in the IID setting, but fails under Holdout shift, leading to the drastic gap in performance.
\begin{figure}
\centering
\includegraphics[width=.9\columnwidth]{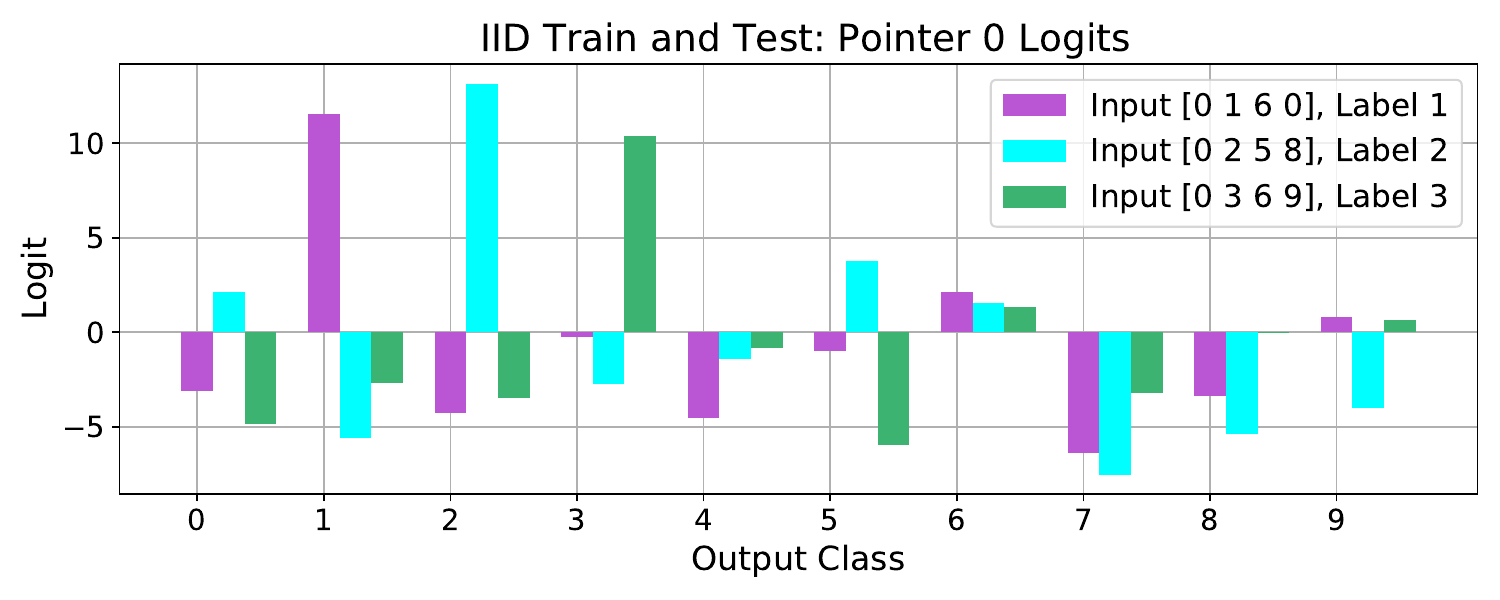} \\
\includegraphics[width=.9\columnwidth]{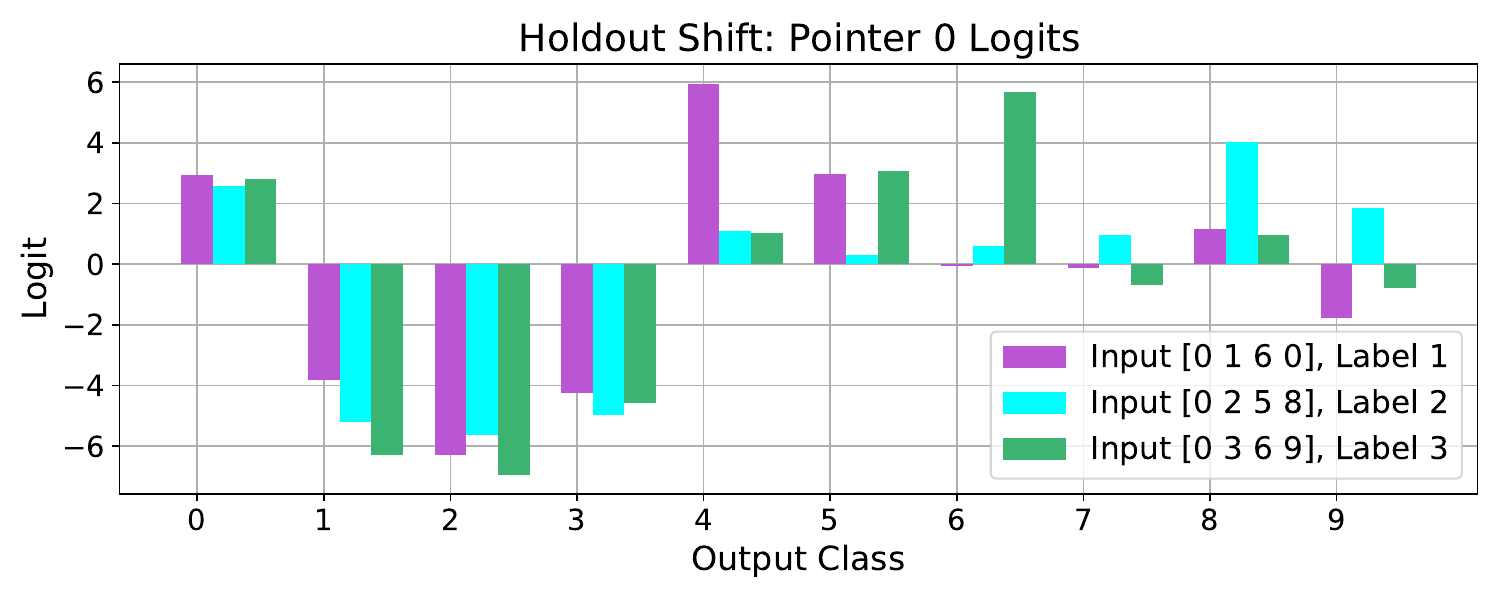}
\vskip-8pt
\caption{\small \textbf{Analyzing the failure of neural networks on Holdout shift reveals systematic mistakes from learned correlations between pointer digits and labels.} We plot the logit values for three examples in the Holdout test set, for a model trained in the IID setting (top) and the Holdout shift setting. The model trained in Holdout shift assigns very low logits to labels $1-3$, because training examples with pointer $0$ are never labelled $1-3$ in the Holdout set. By contrast the IID setting shows no such adversarial correlation. Additional examples are in the Appendix.}
\label{fig:logits-iid-dshift}
\end{figure}

In fact, taking a closer look at test performance in Holdout shift, we see that it converges to \textit{below random} through training, suggesting a learned, systematic error. To understand this further, we take three examples in the Holdout shift test set, and plot the (raw) logit values computed by a model trained in the IID setting vs the Holdout shift setting. These results are shown in Figure \ref{fig:logits-iid-dshift}. 

For these three examples, the pointer is $0$, so the label is the digit value on the top right. In the Holdout test set, this value is always $1-3$. But in Holdout training, there is never a value of $1-3$ on the top right. Thus the networks trained in Holdout shift automatically assign very low logit values for classes $1-3$, while in the IID setting, each of these classes is correctly identified. In summary, under Holdout shift, models learn to \textit{negatively correlate the pointer digit with with values held out during training}. 

This result raise interesting open questions that can be explored on this benchmark. Are there architectural changes that can enforce better priors and withstand distribution shift? Can novel learning objectives prevent these adversarial correlations? Progress on these questions could enable greater robustness in neural network reasoning. 

The result also demonstrates the nuances in trying to understand mechanisms of neural network reasoning. While the network did not learn the pointer rule, its systemic failure arose from spotting a true co-occurrences pattern in the training data. In Section \ref{sec:concepts-and-reasoning}, we return to this consideration, combining distribution shift and functional complexity to further study neural network reasoning capabilities.

\section{Empirical Exploration of PVR Tasks}
\label{sec:warmpu-dshift}
\vspace{-0.4em}
Informed by the preliminary results on distribution shift, we now perform a larger scale study of different PVR tasks, varying model architectures, dataset sizes and task complexities. We observe significant variations in performance, further demonstrating the richness of the PVR testbed, and identify relationships between task complexity and dataset size, with ramifications for neural network reasoning.

\subsection{Experimental Setup}
\vspace{-0.3em}
We focus on token sequence PVR tasks, of length $11$, with each entry being a number in $0-9$ and the left most number acting as a pointer. We vary the functional complexity of these tasks by increasing the window size, termed \textit{number of neighbors} in the experiments. Informed by the study of noise sensitivity (Figure \ref{fig:vec-noise-sensitivity}) which show the \modsum function has a clear monotonic relationship with neighborhood size, we use \modsum to be the aggregation function.

\subsection{Varying dataset size and functional complexity}
\begin{figure}
    \centering
    \includegraphics[width=\linewidth]{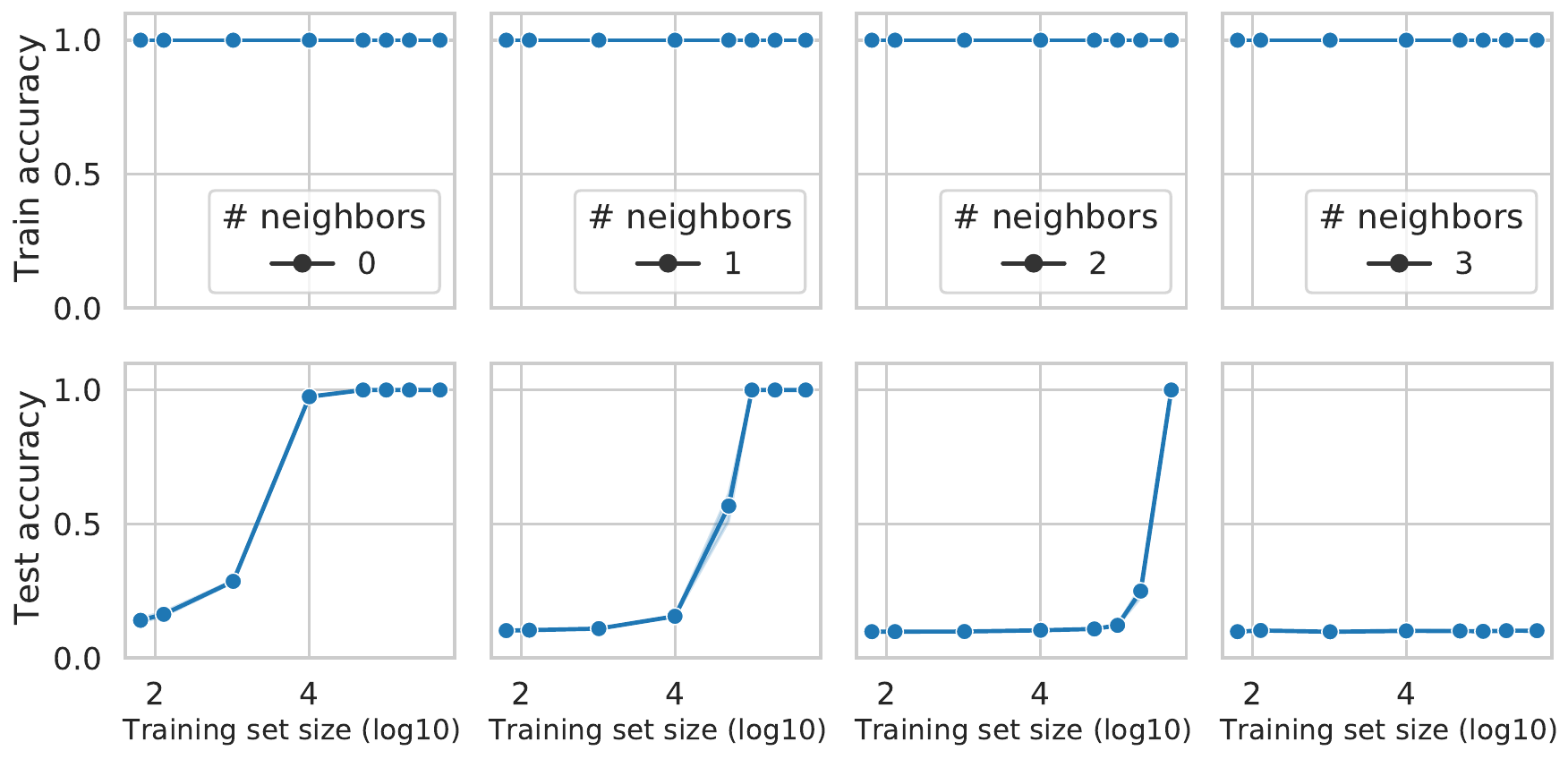}
    \vskip-8pt
    \caption{\textbf{Performance evaluation on PVR tasks when increasing functional complexity and dataset size.} We train \mlp models on vector input PVR tasks of varying complexity (neighborhood size) and training set sizes. While training accuracy is consistently $100\%$, test accuracy starts off random, only showing sharp increases with increased training set size, revealing these models overfit completely at small training set sizes.}
    \label{fig:vec-overview}
\end{figure}

\vspace{-0.3em}
We start by training an \mlp neural network on these token sequence PVR tasks of varying complexity (neighborhood size $0$ to neighborhood size $3$). We also vary the training dataset size going from $\sim10^2$ to $\sim10^6$ datapoints. The results of these experiments are shown in Figure~\ref{fig:vec-overview}, which offers some striking takeaways.

We observe that training accuracy across all complexities (neighborhood sizes) and dataset sizes is consistently $100\%$. This contrasts with test accuracy, which starts off random at lower training set sizes and shows a thresholding behavior, jumping up to $100\%$ with large enough training data. This reveals that at small dataset sizes, the neural network completely overfits, and is not learning the pointer rule. With a larger amount of training data, the test accuracy increases significantly. In Section \ref{sec:concepts-and-reasoning}, we dive further into this, exploring whether $100\%$ test accuracy necessarily means the neural network is learning to reason.  

\begin{figure}[]
    \centering
        \includegraphics[width=.6\linewidth]{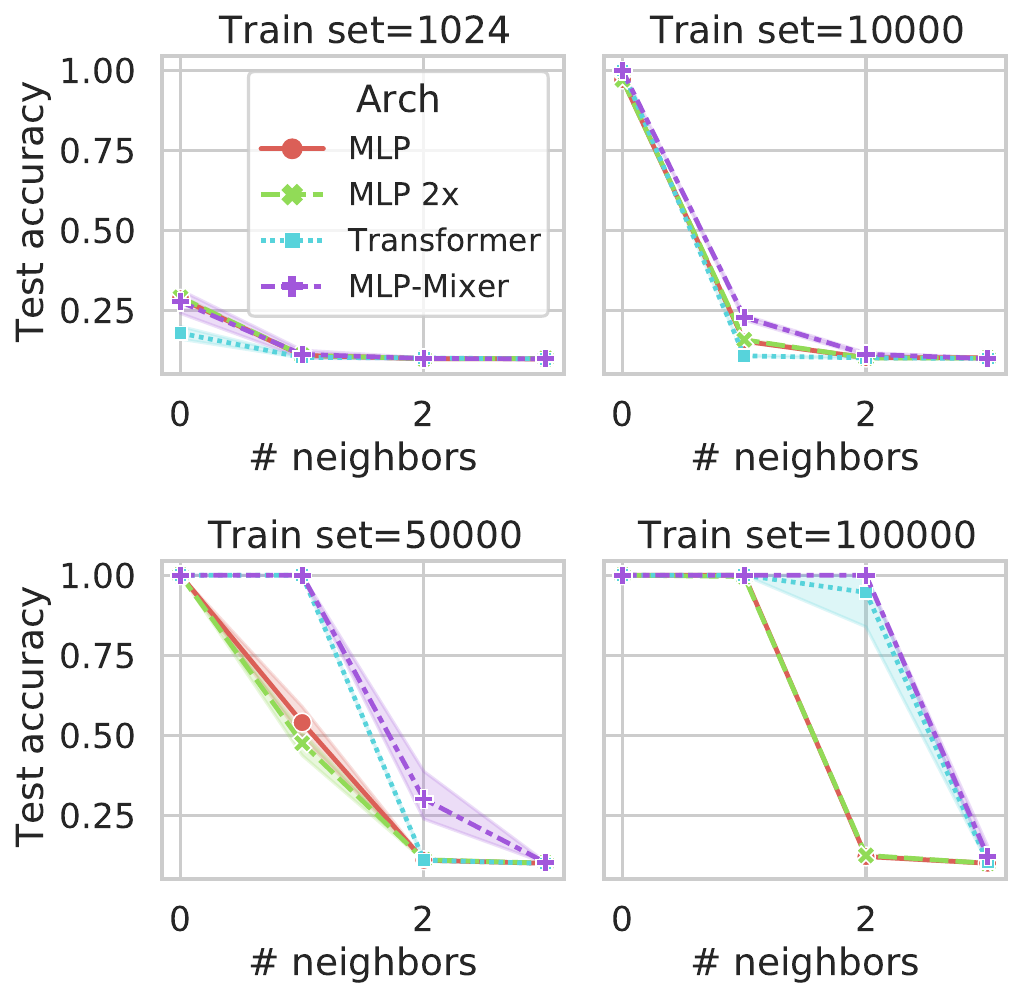}
    \vskip-8pt
    \caption{\textbf{Evaluating vectorized PVR tasks of different complexities demonstrates that \transformer and particularly \mlpmixer have stronger performance.} We show test performance for these different architectures across different training set sizes and functional complexities, finding that \transformer and \mlpmixer show significantly stronger performance, requiring fewer samples for more complex tasks, despite having more parameters. This suggests they have helpful inductive biases.}
    \label{fig:pvr-arch-sweep}
\end{figure}

\subsection{Varying Neural Network Architectures}
\vspace{-0.3em}
We next look at the performance of different neural network architectures on these PVR tasks. The sequential structure of these tasks suggests utilizing alternate models such as the Transformer \citep{vaswani2017attention} and even the recently proposed MLP-Mixer \citep{tolstikhin2021mlp}. In Figure~\ref{fig:pvr-arch-sweep} we show test performances of different architectures as training dataset size and particularly complexity are increased. We observe that \mlp and \mlptwox have similar performance, while the \transformer and particularly \mlpmixer demonstrate significantly better performance --- solving tasks of higher complexity with much less data. These architectures also have larger number of parameters than \mlp (Appendix Section \ref{sec:app-dataset-arch}), so their better sample complexity arises from better inductive biases. The results suggests a future direction of comparing task representations across these architectures to better understand these differences.     

\begin{figure}
    \centering
                                \begin{overpic}[width=.67\linewidth]{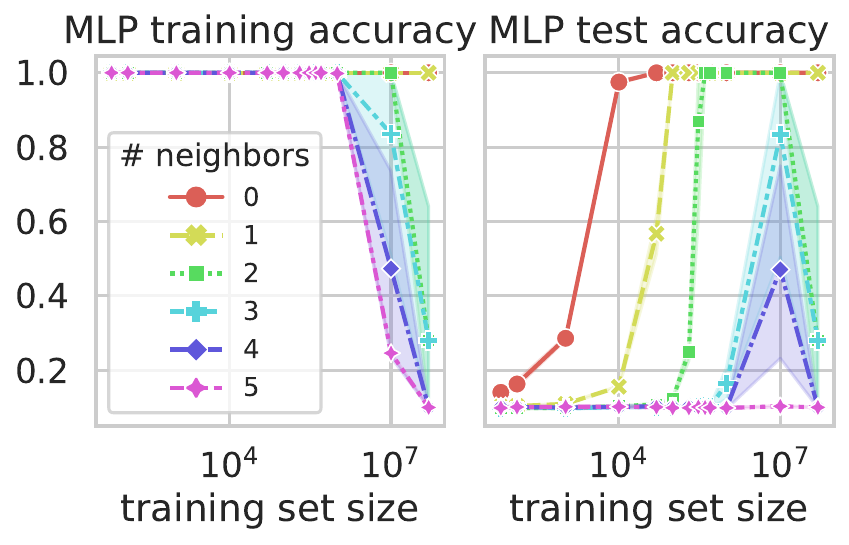}
    \put(1,1){\textbf{(a)}}
    \put(51,1){\textbf{(b)}}
    \end{overpic}
    \begin{Overpic}[width=.31\linewidth]{    \adjincludegraphics[width=.31\linewidth,trim={{.54\width} 0 0 0},clip]{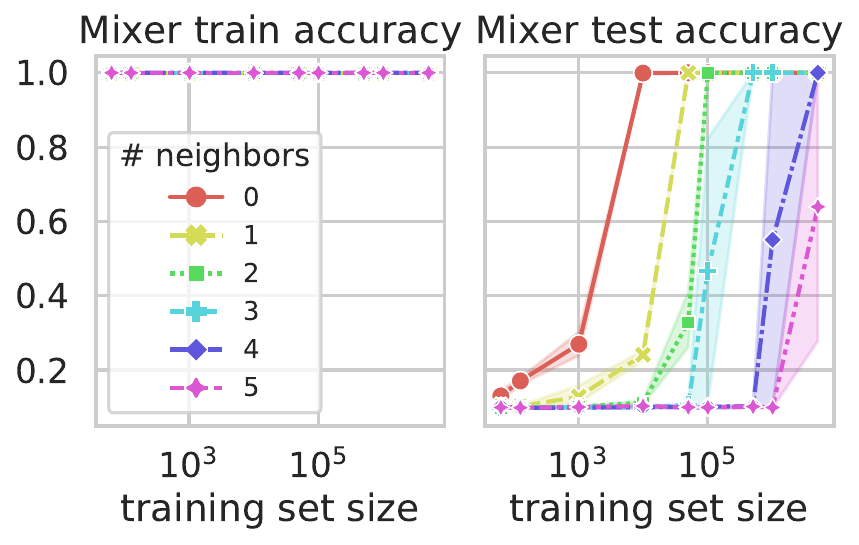}    }\put(-4,1){\textbf{(c)}}\end{Overpic}
    \vskip-8pt
    \caption{\textbf{Training (a) and test (b) performance \mlp, test performance of (c) \mlpmixer at massive training set sizes.} Both models show ability to tackle more complex tasks with increased training data, though \mlp saturates on capacity, struggling to fit the training dataset at $10^7$ datapoints. \mlpmixer sees consistent improvements and can solve tasks of complexity $m$ at $\sim 5\times 10^{m+3}$ datapoints, and could achieve perfect training accuracy (not shown) for all cases tested here. Though \mlpmixer runs show greater variance on more complex tasks: sometimes the optimization fails. We find that at high complexities some runs \textit{fail to learn}, and we ignore runs where the \emph{training} accuracy is below 20\%.}
    \label{fig:tr-sizes}
\end{figure}

\subsection{Scaling to Massive Dataset Sizes}
Having seen the generalization jumps from increasing training dataset and better neural network architectures, we look at scaling datasets by another order of magnitude, up to $5\times 10^7$ which we use to train different architectures. We show the results for \mlp and \mlpmixer in Figure \ref{fig:tr-sizes}. For \mlp, we observe that with ~$10^7$ training points, it is able to learn neighborhoods of size $3$ (which was at random accuracy in the smaller sweep of Figure \ref{fig:vec-overview}), but at larger number of datapoints, struggles to fit the training set. \mlpmixer, with its larger capacity and better inductive bias, shows continuing performance improvements as dataset size is increased, solving more and more complex tasks, although we see variance increase with the most complex tasks, with \textit{some seeds failing to train} (suggesting open questions on learning dynamics). For \mlpmixer, we see $100\%$ test accuracy with ~$5\times 10^{m+3}$ data for tasks of complexity $m$

\section{Fast vs Slow Learning, and Reasoning}
\vspace{-0.8em}
The previous experiments across different architectures, dataset sizes and task complexities reveal two phenomena on learning and reasoning: (i) learning high complexity PVR tasks is unstable, with failed runs (Figure \ref{fig:tr-sizes}) (ii) good test accuracy on higher task complexity requires larger training data (Figures \ref{fig:vec-overview}, \ref{fig:tr-sizes}). We study these below, uncovering surprising insights about neural network reasoning.   
\subsection{Fast Learning vs Slow Learning}
\vspace{-0.3em}
\begin{figure}
    \centering
    \includegraphics[width=.47\linewidth]{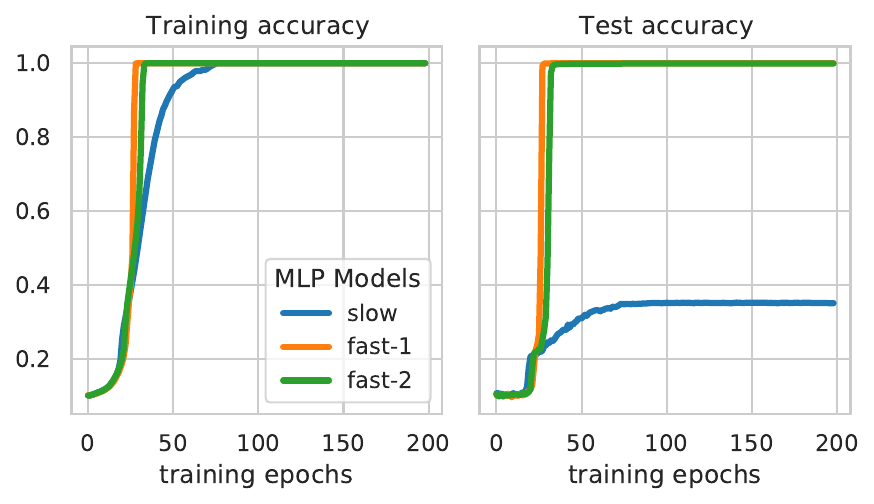}
    \hfill
    \includegraphics[width=.47\linewidth]{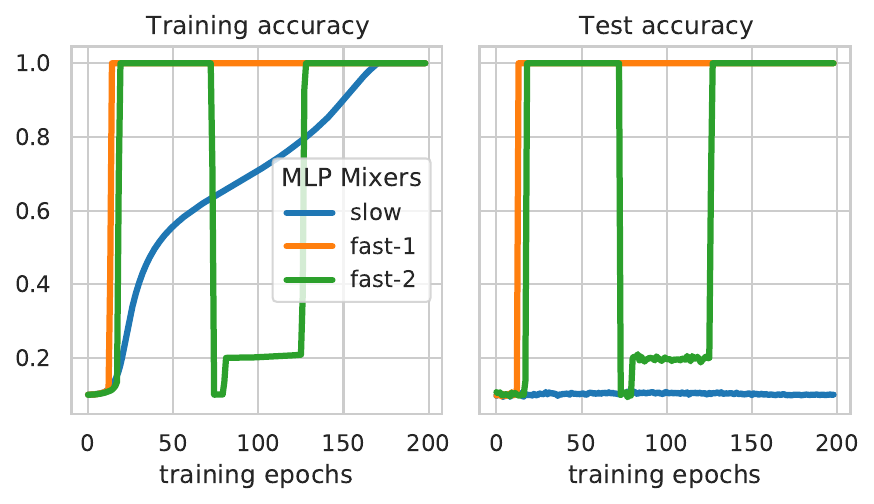}
    \vskip-8pt
    \caption{\small \textbf{Fast vs Slow learning in \mlp and \mlpmixer.} Learning curves for (left) MLPs and (right) MLP-Mixers that depict fast learners that generalize well and slow learners that generalize poorly. The MLPs are trained on 300k training examples with 2 neighbors. The MLP-Mixers are trained on 5M training examples with 5 neighbors.}
    \label{fig:fast-vs-slow-body}
\end{figure}
\begin{figure}
    \centering
    \includegraphics[width=.5\linewidth]{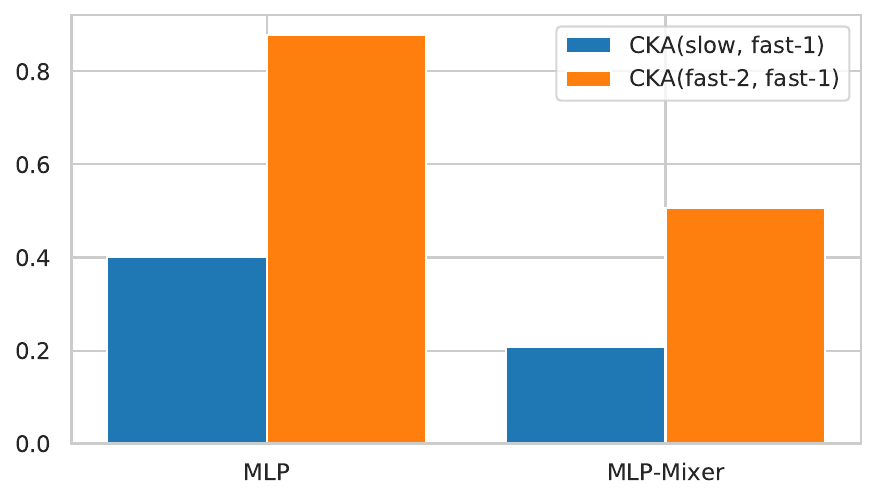}
    \vskip-10pt
    \caption{\textbf{Two independent fast learners have more similar representations than a fast learner and a slow learner.} The centered kernel alignment (CKA) similarity between the two pairs of models from Figure~\ref{fig:fast-vs-slow-body}.}
    \label{fig:fast-vs-slow-cka}
\end{figure}

Inspecting the runs that: (1) fail to learn, or (2) (more interestingly) get 100\% training accuracy but fail to generalize, we find that failing runs typically converge \textit{slowly}, while successful runs learn quickly.
Although both modes reaches 100\% training accuracy, internally they might be learning very different representations, only one of which generalizes well. To test this hypothesis, we take the two fast learners and the slow learners shown in Figure~\ref{fig:fast-vs-slow-body}, and then compute how similar their learned representations are (on the training dataset of the slow learner), using Centered Kernel Alignment \citep[CKA,][]{kornblith2019similarity}. 
As shown in Figure~\ref{fig:fast-vs-slow-cka}, for both MLP and MLP-Mixer cases, the CKA similarity between two fast learners are higher than slow-vs-fast. This suggests that two independent fast learners are learning representations that are more similar to each other than to an independent slow learner, even when measured on the training data of the slow learner, where it achieves 100\% accuracy. There remain rich open questions in exploring what causes fast or slow learning to arise. 

\subsection{Training Data, Test Accuracy and Reasoning}
\label{sec:concepts-and-reasoning}
\vspace{-0.4em}
The need for larger training data with more complex tasks also raises key question: is the high test accuracy indicative of learning reasoning, or for a task of complexity $m$, are the models simply \textit{memorizing} the $10^{m+2}$ digits corresponding to all the different choices of (i) pointer (ii) neighborhood digits (iii) the associated label? To answer this, we look at performance when extending Holdout shift to higher functional complexities.

\begin{figure}
    \centering
    \adjincludegraphics[width=.48\linewidth,trim={{.55\width} 0 0 0},clip]{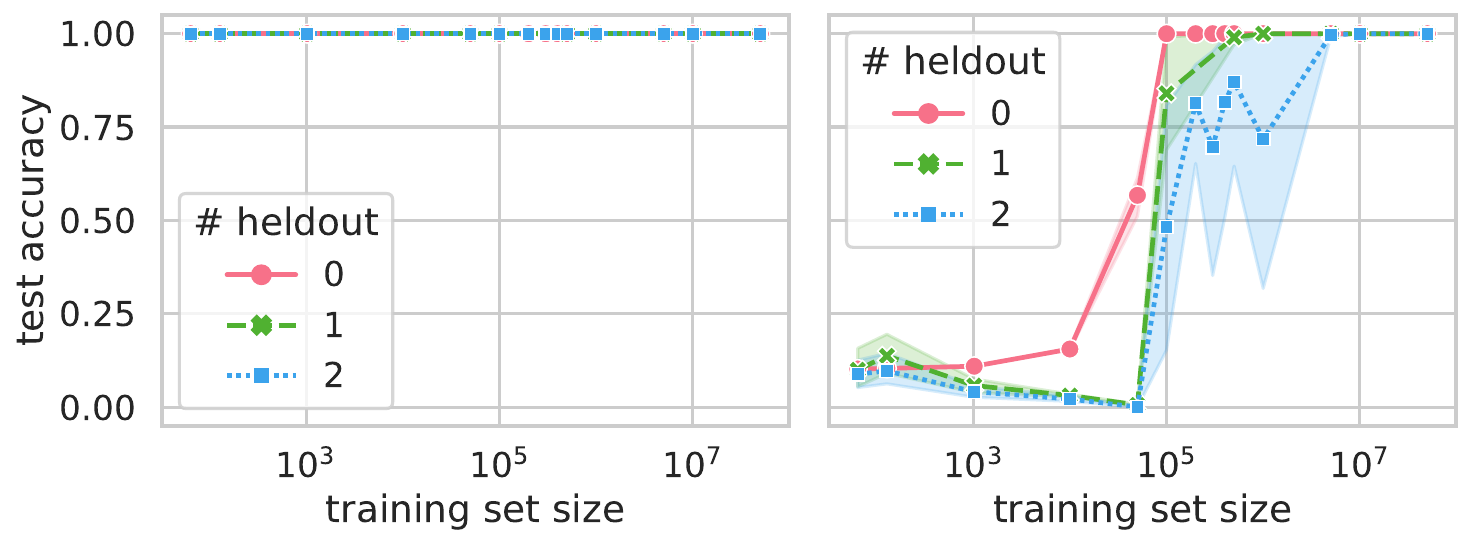}
    \adjincludegraphics[width=.48\linewidth,trim={{.55\width} 0 0 0},clip]{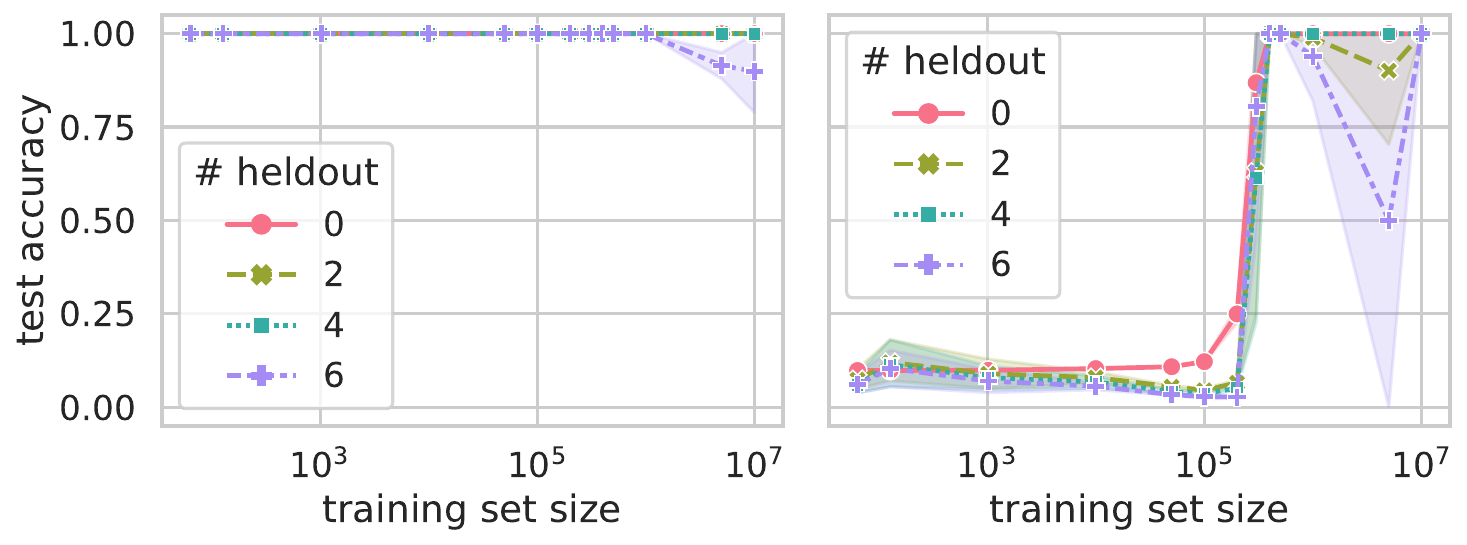}
        \begin{tabular}{p{.4\linewidth}p{.4\linewidth}}
    \textbf{(a)} window size 2 & \textbf{(b)} window size 3
    \end{tabular}
    \vskip-7pt
    \caption{\small \textbf{A holdout shift experiment at higher complexity shows neural networks can generalize to truly unseen instances --- if they don't fail at training}. We train neural networks on task complexities $m=1,2$ with up to $(m+1)!$ permutations of $(0,1,\ldots,m)$ heldout from the window of values at training, and test on \textit{only} examples with $(0,1,\ldots,m)$ in the values window. For $m=3$, training is often unstable, and we discard runs with training accuracy less than $60\%$. Strikingly, when training succeeds, the models achieve $100\%$ accuracy on the holdout test set, strongly suggesting they are learning to reason.}
    \label{fig:holdout-many}
\end{figure}
Specifically, for a task of complexity $m$, we ensure the sequence $(0,1,\ldots,m)$ never occurs in the value window in the training set. Meanwhile, the test set is adversarially constructed to contain only $(0,1,\ldots,m)$ in the value window, but the pointers and other digits outside the value window are random. We call this task \texttt{holdout-1}. Similarly, \texttt{holdout-2},...,\texttt{holdout-i}, etc. are constructed to hold out not only $(0,1,\ldots,m)$, but also $i$ permutations of it. 

We train neural networks on this holdout task for $m=1,2$, holding out up to $(m+1)!$ permutations. Like with tasks of high complexity, there are slow learners that fail to learn well, shown in Appendix Figure \ref{fig:fast-vs-slow-holdout}. But if we exclude these runs, the results, in Figure \ref{fig:holdout-many}, illustrate a remarkable success --- the neural networks generalize even when \textit{all} permutations are held out. Together, these results suggest that neural networks do have sophisticated reasoning capabilities, but the structure of the task, inductive bias of the models and properties of the data can all have highly influential roles. There remain many exciting open directions on the PVR benchmark in better understanding this interplay and its effects on learning and reasoning.    \section{Discussion}
\label{sec:discussion}
Understanding the mechanisms of artificial neural network generalization is both of fundamental scientific importance and key to their robust uses in many high-stakes settings.  As steps towards this, we propose a new benchmark, Pointer Value Retrieval (PVR), to explore the limits of artificial neural network reasoning. Informed by studies in cognitive neuroscience, which show that the capability to process \textit{indirection} serves as a cornerstone for human reasoning, PVR incorporates indirection in the form of pointers referencing positions and values. This structure allows PVR tasks to have a diversity of possible input modalities (incorporating aspects of representation learning) and have systematically varying task complexity through increased functional complexity and distribution shift. An empirical investigation across different PVR tasks reveals many insights on neural network generalization, from the role of inductive biases in different architectures to a propensity to rely on nearest neighbor styled mechanisms to a tight connection between task complexity and training dataset size. Further investigation of high complexity PVR tasks uncovers key differences between fast and slow learners, and partial successes for sophisticated reasoning in the presence of significant distribution shift. These findings raise many open questions to explore on this benchmark, from variations of representations across architectures to the dynamics of the learning process, all which offer promising future progress in understanding generalization and reasoning capabilities of artificial neural networks.

\bibliographystyle{apalike}
\bibliography{refs}

\appendix\clearpage\onecolumn

\section{Dataset and Architecture Details}
\label{sec:app-dataset-arch}
\textbf{PVR Datasets:} In the main text, we perform experiments on:
\begin{itemize}
    \item \textbf{visual block style MNIST PVR IID dataset} For this dataset, which is used in Section \ref{sec:warmpu-dshift}, for the training set, we sample each position iid from the $60000$ MNIST training examples. For the testset, we similarly sample each position iid from the $10000$ MNIST test set. To introduce some more variation, each block position is of size $40$x$40$, and we randomly jitter the $28$x$28$ MNIST digit within the block.
    \item \textbf{visual block style MNIST PVR Holdout Shift dataset} Here for the training set, we again sample from the MNIST training set, but holding out some digits at some positions. So for the top right, we don't sample from digits $1-3$ (but sample iid from the remaining MNIST digits), for the bottom left, we don't sample from $4-6$ (but iid from all other MNIST training digits) and for the bottom right we don't sample from MNIST digits that are $7-9, 0$. The pointer is sampled iid from all MNIST training digits. For the test set we sample from the MNIST test digits, but again holding out some digits, i.e. we \textit{only} sample iid from MNIST test digits $1-3$ for the top right corner, and so on. We also jitter each digit like above.
    \item \textbf{vectorized PVR dataset} For the vectorized PVR dataset, we simply sample iid from $0-9$ for each position in the sequence.
\end{itemize}

For the visual PVR tasks, specifically the MNIST visual PVR task, we use the standard \resnet and \vgg, from the \texttt{torchvision} models library. We train for $10$ epochs with batch size $50$. We use the Adam optimizer with learning rate $10^{-4}$.
 
For the vectorized PVR tasks, we use the following network architectures. Table~\ref{tab:model-sizes} summarize the number of parameters in each model we use.
\begin{itemize}
    \item \mlp: We use an embedding layer with vocabulary size 10 and embedding dimension 64 to map each of the input token to a vector representation. The concatenated representations from all input tokens are then passed through 4 fully connected layers with output dimension 512, 1024, 512, and 64, respectively. We apply ReLU activation function after each fully connected layer. Finally, a linear layer with output dimension 10 is attached as the classifier.
    \item \mlp $2\times$: The same as \mlp, except that the output dimensions of the 4 fully connected layers are doubled: 1024, 2048, 1024 and 128.
    \item \transformer: We use the encoder part of standard Transformer~\citep{vaswani2017attention}, which consists of multiple \emph{transformer layer}, where each transformer layer consists of a multi-head self-attention block and a MLP block. In particular, we use embedding dimension 512, 4 transformer layers, with 4 heads for each self-attention block and hidden dimension 1024 for each MLP block. Following~\citet{dosovitskiy2020image}, we prepend a virtual \emph{class token} with learnable vector representations, and in the final encoder output, attach a linear classifier to the representation of that token for classification.
    \item \mlpmixer: MLP-Mixer is a new architecture recently proposed by \citet{tolstikhin2021mlp}. It is similar to Transformers, except that the multi-headed self-attention layers are replaced with fully connected layers. We use embedding dimension 512 and 4 \emph{mixer layers}, with token-MLP dimension 768 and channel-MLP dimension 2048. Similar to \transformer, we use a \emph{class token} for the purpose of classification.
\end{itemize}

We did a small scale hyperparameter sweep on different optimizers (SGD, Adam~\citep{kingma2014adam}, Adagrad~\citep{duchi2011adaptive}, LAMB~\citep{you2019large}), learning rates ($1\times 10^{-3} \sim 4\times 10^{-1}$) and batch sizes (128, 256, 512, 1024, 2048, 4096). In the end we choose the following hyperparameters by balancing performance and stability across different setups: We use weight decay $1\times 10^{-5}$, SGD optimizer with momentum 0.9, and cosine learning rate scheduling with base learning rate 0.05, and a linear warmup period of 10 epochs. The batch size is 1024, and we train for 200 epochs. For studies with tiny training sets (e.g. 64), we train for at least 800 iterations. The training was done with NVidia P100 / V100 GPUs.

\begin{table}[]
    \centering
    \caption{Number of parameters for models used in the empirical studies.}
    \label{tab:model-sizes}
    \begin{tabular}{cc}\toprule
    Architecture & Parameter count \\\midrule
    \mlp & 1,445,194  \\
    \mlp $2\times$ & 5,052,426 \\
    \mlpmixer & 8,495,674 \\
    \transformer & 8,429,066 \\
    \bottomrule
    \end{tabular}
\end{table}

\section{Additional Results from Initial Experiments on Visual PVR and Distribution Shift}
\begin{figure}[h]
\centering
\begin{tabular}{cc}
    \includegraphics[width=0.5\columnwidth]{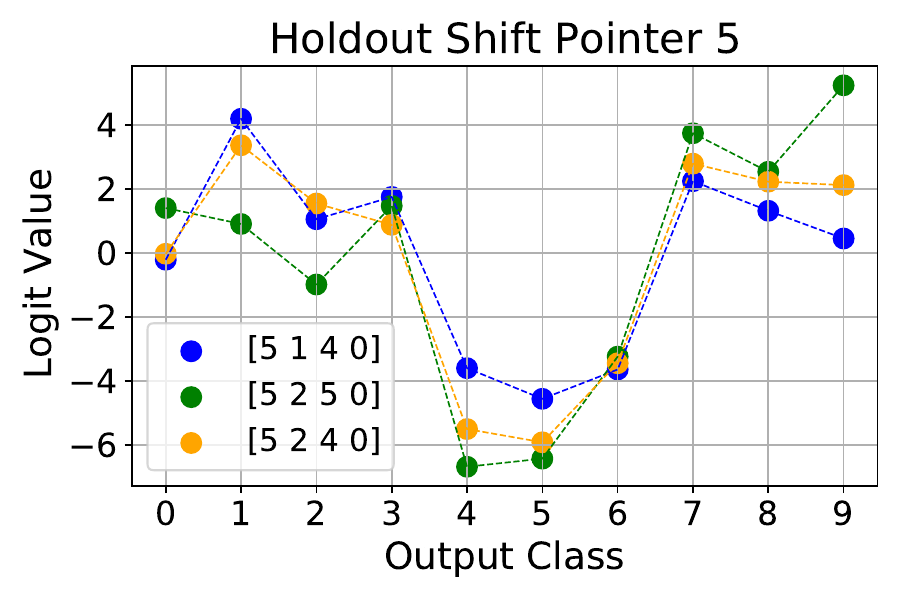} &
    \includegraphics[width=0.5\columnwidth]{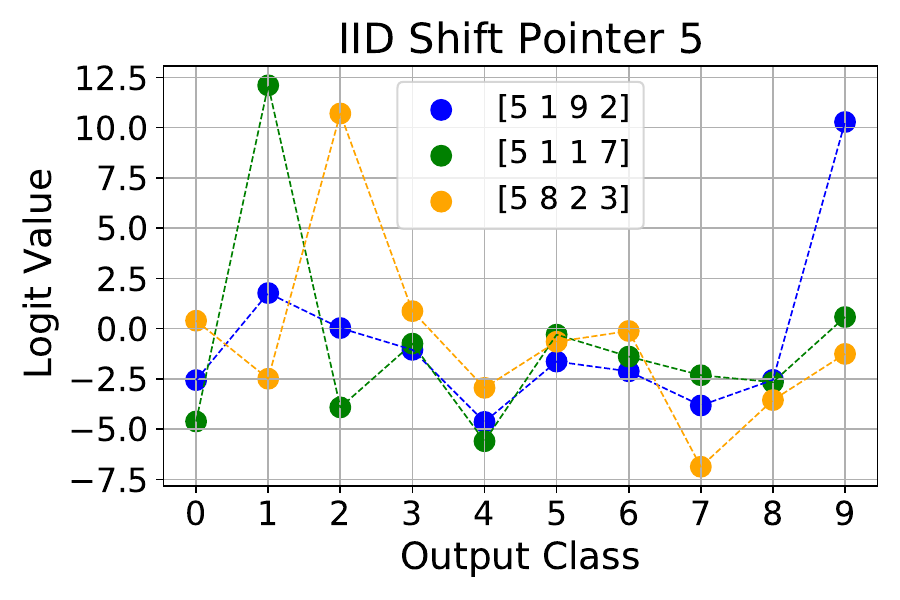} \\
    \includegraphics[width=0.5\columnwidth]{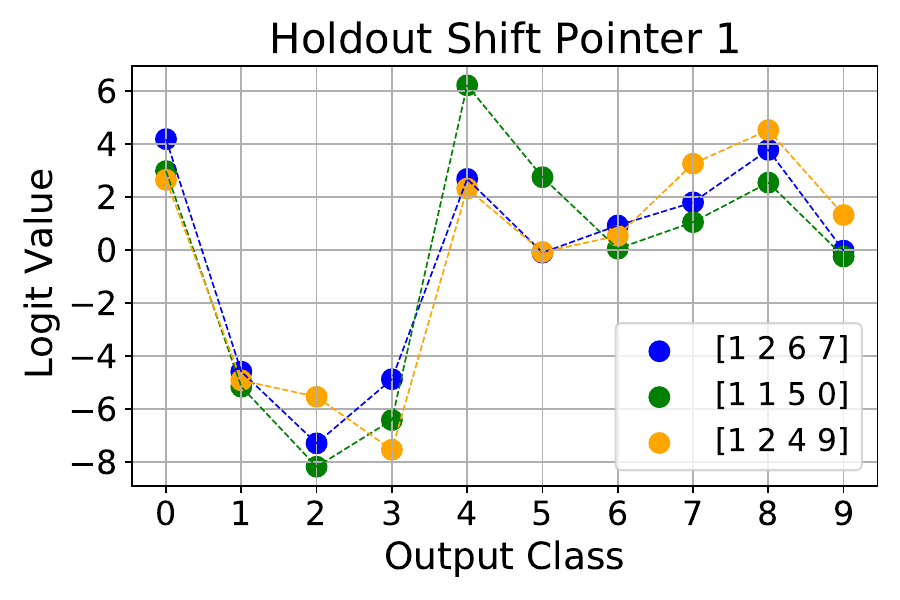} &
    \includegraphics[width=0.5\columnwidth]{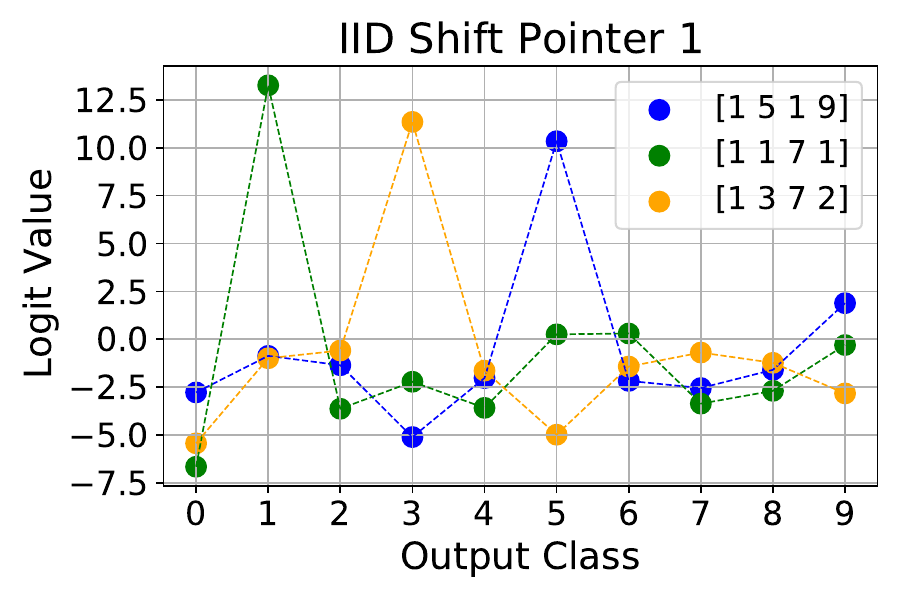}
\end{tabular}
\caption{\small \textbf{Analyzing the failure of neural networks on Holdout shift reveals systematic mistakes from  learned correlations between pointer digits and labels.} Additional logit results, compare to Figure \ref{fig:visual-dshift-schematic} in the main text.}
\label{fig:visual-dshift-schematic-app}
\end{figure}

Additional results can be found in Figure~\ref{fig:visual-dshift-schematic-app}.

\section{Noise Sensitivity Analysis}

We applied \emph{noise sensitivity} of boolean functions in the main text to characterize the complexity of our vectorized PVR tasks with different neighborhood sizes and aggregation functions. Specifically, for a function $f$ with boolean (binary bits) inputs, and $\delta\in[0,1]$, the \emph{noise sensitivity} $\nsmath_\delta[f]$ measures the probability that $f(x)\neq f(y)$, where $x$ consists of uniformly random bits, and $y$ is formed from $x$ by flipping each bit independently with probability $\delta$. It is widely used to measure the stability and closely related to various complexity measure of boolean functions. For example, for $\delta\in (0, 1/2]$, the Fourier spectrum of a binary output boolean function $f$ can be shown to be $\epsilon$-concentrated on degree up to $1/\delta$ with $\epsilon\leq 3\nsmath_\delta[f]$~\citep[][Proposition~3.3]{o2014analysis}. 

We extend it to measure the complexity of our target functions by representing the inputs of our vectorized PVR task as bit vectors. In particular, each digit $x\in\{0,\ldots,K-1\}$ in our input sequence is represented as $D$ bits by the standard binary representation of unsigned integers, where $D$ is the smallest integer such that $K\leq 2^D$. When $K\neq 2^D$, a subset of random uniform bit sequences will fall out of the valid range of $x$. We simply extend the definition of the target functions to take digits with arbitrary values, but convert them into the ``valid range'' via $\text{mod } K$ as a preprocessing. 

\begin{figure}
    \centering
                \includegraphics[width=.9\linewidth]{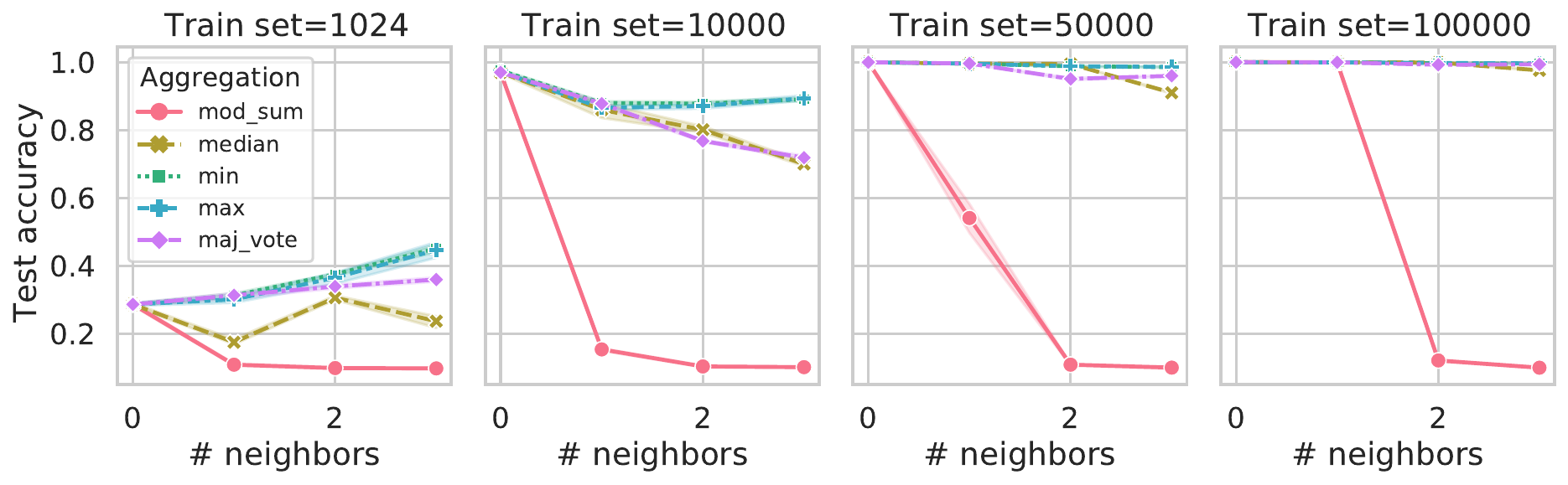}
    \caption{\small \textbf{Evaluating different aggregation functions for PVR tasks shows \modsum is the most challenging.} We show test performance for different aggregation functions across varying dataset size and functional complexity. The empirical results support the intuitive observation that \modsum is the most challenging.}
    \label{fig:pvr-aggr-sweep}
\end{figure}
Figure~\ref{fig:pvr-aggr-sweep} illustrates test accuracy for different aggregation functions across varying PVR functional complexity and dataset sizes. The learning results are consistent with our measurements of task complexity using noise sensitivity: the tasks become noticeably more challenging to learn as the neighborhood size increases only when using the \modsum aggregation function.

\section{PVR without Pointers}

\begin{figure}
    \centering
    \begin{subfigure}[b]{\textwidth}
         \centering
         \includegraphics[width=\linewidth]{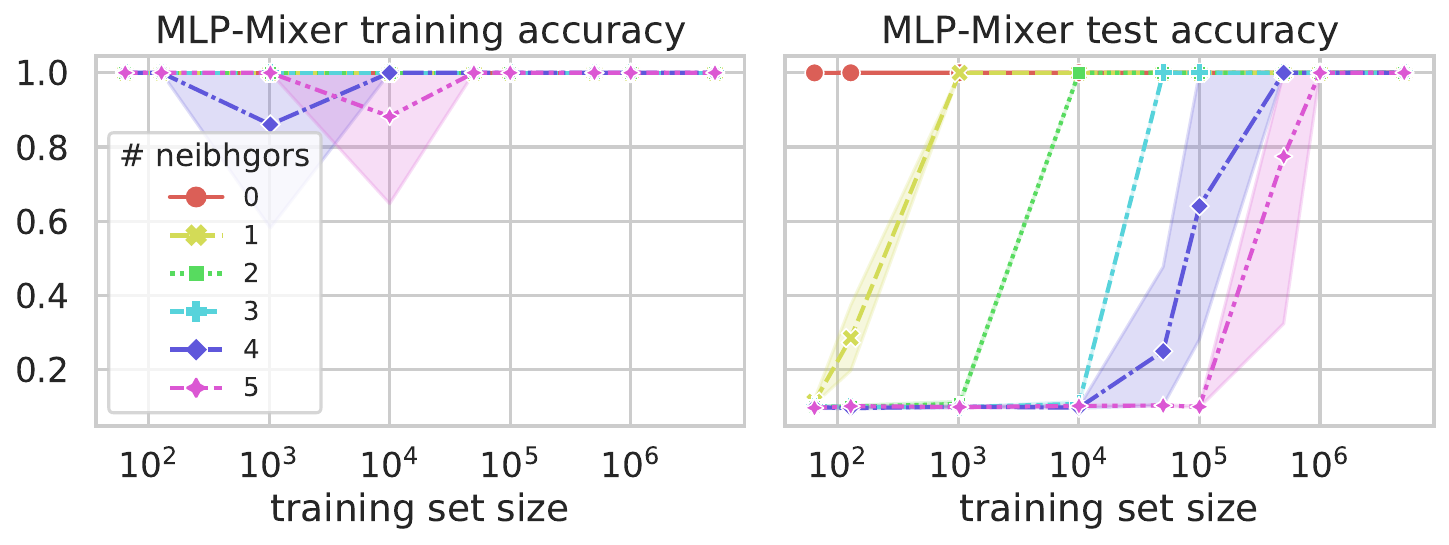}
         \caption{Without pointer}
     \end{subfigure}
    \begin{subfigure}[b]{\textwidth}
         \centering
         \includegraphics[width=\linewidth]{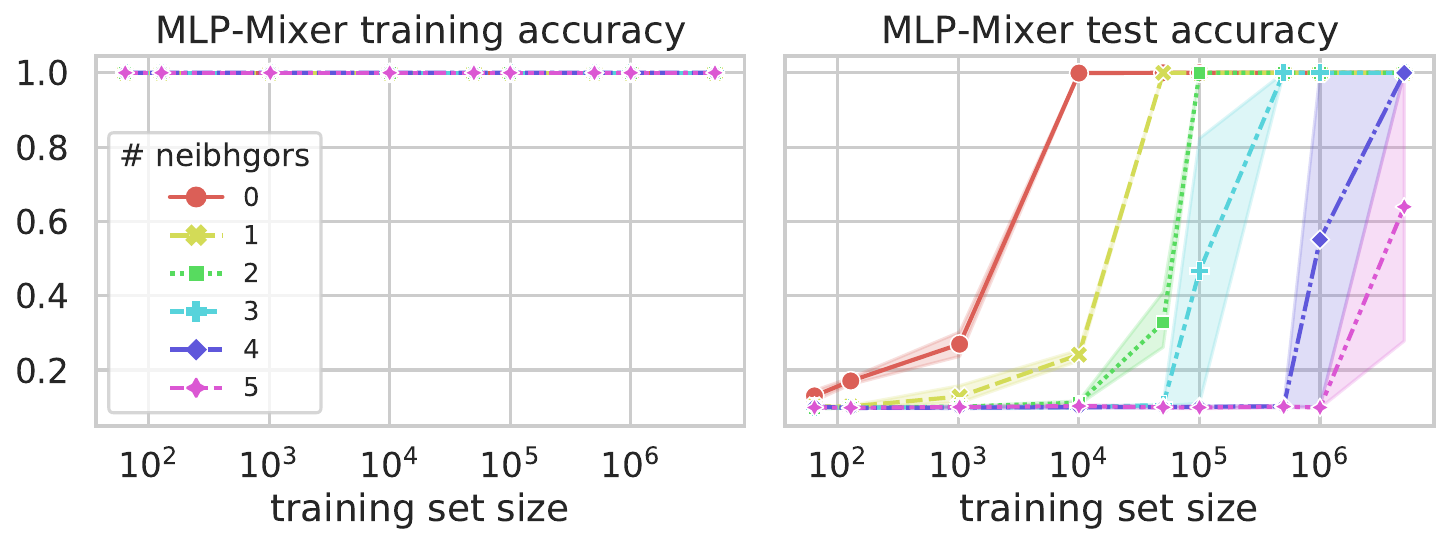}
         \caption{With pointer}
     \end{subfigure}
    \caption{\textbf{Comparing performance evaluation on PVR tasks with and without pointers.}}
    \label{fig:vec-overview-fixed-pointer0}
\end{figure}

The PVR tasks consists of a pointer and a value aggregation function in a window. In this section, we study the role of the pointer by ``removing'' the pointer feature from the PVR tasks. In particular, we force the data generation procedure to always use 0 as the pointer, and then run the same training setup as in the case of standard PVR tasks. Figure~\ref{fig:vec-overview-fixed-pointer0} compares the MLP-Mixer performances on PVR tasks with and without pointers. As expected, when a pointer is not used, the tasks are much easier and could require several order of magnitudes fewer training examples in order to learn.

\section{Fast Learning vs Slow Learning with Holdout}

\begin{figure}
    \centering
    \includegraphics[width=.6\linewidth]{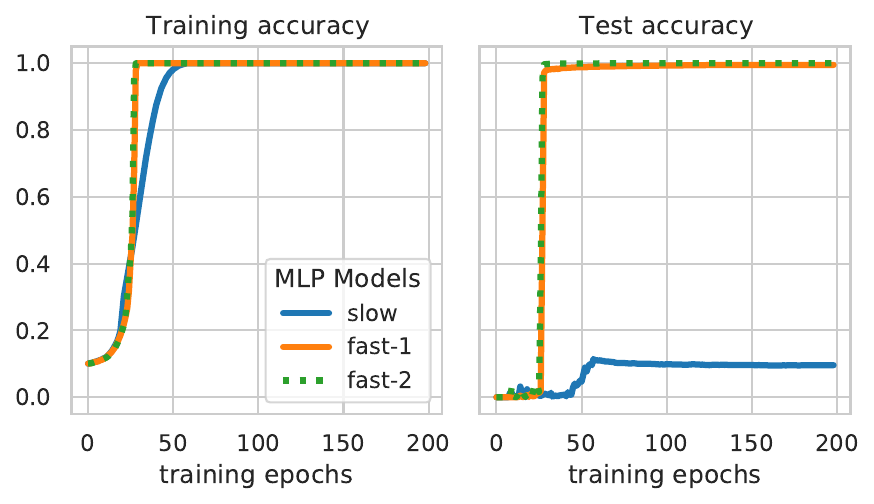}
    \caption{Learning curves for MLPs trained with holdout that depict fast learners that generalize well and slow learners that generalize poorly. The models are trained on 300k training examples of the PVR task with window size 3, and all 6 permutations of \{1, 2, 3\} in the value window being holdout from the training set.}
    \label{fig:fast-vs-slow-holdout}
\end{figure}

Figure~\ref{fig:fast-vs-slow-holdout} shows the fast learning vs slow learning phenomenon for the case with holdout data. Following the analysis in the main text, we use CKA to estimate the similarity between learned representations, and find that the representations between fast-1 and fast-2 are more similar (CKA=0.9146) than between fast-1 and slow (CKA=0.3182).

\end{document}